\documentclass[nonacm,sigconf]{acmart}

\usepackage[utf8]{inputenc}
\usepackage[english]{babel}
\usepackage{amsmath}
\usepackage{multirow,multicol,booktabs,makecell,array}
\usepackage[
  separate-uncertainty=true,
  table-align-uncertainty=true,
  detect-all
]{siunitx}
\usepackage{parskip}
\usepackage{tablefootnote}
\usepackage{hyperref}
\usepackage{adjustbox}
\usepackage{float}
\usepackage{paralist}
\usepackage[inline]{enumitem}
\usepackage{xcolor}
\usepackage{xspace}
\usepackage{bm}

\usepackage[capitalize]{cleveref}
\crefname{section}{Sec.}{Secs.}
\Crefname{section}{Section}{Sections}
\Crefname{table}{Table}{Tables}
\crefname{table}{Tab.}{Tabs.}

\newcommand{\CI}{\texttt{CI}\xspace}
\newcommand{\HV}{\texttt{HV}\xspace}

\newcommand{\PBC}{PBC\xspace}
\newcommand{\SPP}{SPP\xspace}
\newcommand{\FRM}{FRM\xspace}
\newcommand{\BCM}{BCM\xspace}
\newcommand{\BCR}{BCR\xspace}

\AtBeginDocument{%
  }

\setcopyright{acmlicensed}
\copyrightyear{2025}
\acmYear{2025}
\acmDOI{XXXXXXX.XXXXXXX}

\acmConference[Conference acronym 'XX]{Make sure to enter the correct
  conference title from your rights confirmation email}{June 03--05,
  2018}{Woodstock, NY}

\acmISBN{978-1-4503-XXXX-X/18/06}

\begin{document} 

\title{Interpretable Non-linear Survival Analysis with\\Evolutionary Symbolic Regression}


\author{Luigi Rovito}
\orcid{0000-0003-2772-4095}
\affiliation{%
  \institution{University of Trieste, Italy}
  \country{}
}
\email{luigi.rovito@phd.units.it}

\author{Marco Virgolin}
\orcid{0000-0001-8905-9313}
\affiliation{%
  \institution{InSilicoTrials Technologies, The Netherlands}
  \country{}
}
\email{marco.virgolin@insilicotrials.com}

\renewcommand{\shortauthors}{Rovito et al.}

\begin{abstract} 
Survival Regression (SuR) is a key technique for modeling time to event in important applications such as clinical trials and semiconductor manufacturing.
Currently, SuR algorithms belong to one of three classes: non-linear black-box---allowing adaptability to many datasets but offering limited interpretability (e.g., tree ensembles); linear glass-box---being easier to interpret but limited to modeling only linear interactions (e.g., Cox proportional hazards); and non-linear glass-box---allowing adaptability and interpretability, but empirically found to have several limitations (e.g., explainable boosting machines, survival trees).
In this work, we investigate whether Symbolic Regression (SR), i.e., the automated search of mathematical expressions from data, can lead to non-linear glass-box survival models that are interpretable and accurate.
We propose an evolutionary, multi-objective, and multi-expression implementation of SR adapted to SuR. 
Our empirical results on five real-world datasets show that SR consistently outperforms traditional glass-box methods for SuR in terms of accuracy per number of dimensions in the model, while exhibiting comparable accuracy with black-box methods.
Furthermore, we offer qualitative examples to assess the interpretability potential of SR models for SuR.
Code at: \url{https://github.com/lurovi/SurvivalMultiTree-pyNSGP.}.
\end{abstract}




\maketitle

\emph{This is the author's version of the work. It is posted here for your personal use. Not for redistribution. The definitive Version of Record was published in Proceedings of the Genetic and Evolutionary Computation Conference 2025, http://dx.doi.org/10.1145/3712256.3726446}

\section{Introduction}
\label{sec:intro}

Survival regression (SuR) is a foundational approach for modeling and analyzing time to event data.
In drug development, SuR can lead to insights on the safety and effectiveness of treatments~\cite{collett2015modelling,cox1972regression,GEORGE2014686}.
Events of interest for SuR in such contexts include outcomes such as death or key indicators of disease progression, such as relapses of multiple sclerosis~\cite{koch2023relapse}. 
SuR is also valuable in other fields, such as the manufacturing industry, where it can be used to model the time until the failure or breakdown of machinery components~\cite{hrnjica2021survival}.

SuR modeling approaches can be generally categorized into three classes based on their interpretability and functional form:
\begin{enumerate}
    \item \textbf{Black-box and non-linear}: These models, such as neural networks and random forests, can capture complex relationships in the data~\cite{Wiegrebe2024,Pickett2021}.
    However, their lack of transparency can be met with skepticism among practitioners, e.g., in healthcare~\cite{singh2011survival}.
    
    \item \textbf{Glass-box and linear}: Models in this category, such as the Cox proportional hazards model, use linearity to capture feature relationships.
    Linearity brings ease of interpretation, but can limit predictive accuracy~\cite{Sundrani2021}.
    
    \item \textbf{Glass-box and non-linear}: These models strive to balance complexity and interpretability.
    They are designed to capture non-linear relationships while maintaining a level of transparency that allows some level of interpretation.
    This category includes methods such as generalized additive models~\cite{Bai2017}, explainable boosting machines~\cite{lou2013accurate}, and survival trees~\cite{Bou-Hamad2011}. Despite their potential, the state-of-the-art faces inherent limitations, described below.
\end{enumerate}

Among these three classes, glass-box non-linear algorithms are arguably the most promising~\cite{rudin2019stop}.
Current popular glass-box non-linear algorithms for SuR face several limitations, which we highlight here.
Generalized additive models (GAMs) use univariate (i.e., single-feature) smooth functions called \emph{basis functions}, which are typically realized as polynomials or splines~\cite{Bai2017}. 
The (manual) choice of basis functions can be non-trivial, and greatly influence the accuracy and ease of interpretation the model can achieve.
Explainable boosting machines use gradient boosted trees, which are black-box, but limit them to bivariate interactions to enable plotting and thus interpretation by visualization~\cite{lou2013accurate}.
A limitation of this approach is that the number of visualizations grows with the number of bivariate interactions ($n(n-1)/2$), quickly becoming too large for pragmatic use. 
Lastly, survival trees carry the limitations of decision trees for classification and regression, such as poor generalization due to predicting constant values outside the boundaries of the training data~\cite{bengio2010decision,costa2023recent}.
An alternative worth considering is the use of black-box models paired with explanation methods such as local intepretable model-agnostic explanations~\cite{ribeiro2016should} and Shapley values~\cite{vstrumbelj2014explaining}.
However, explanation methods can only approximate the behavior of the model, and therefore can draw incorrect explanations, and at times even contradict each other~\cite{ghorbani2019interpretation,slack2020fooling,dombrowski2019explanations,adebayo2018sanity,alvarez2018robustness,lee2019developing}.

Stemming from a need to overcome the limitations of the state-of-the-art, this work explores whether Symbolic Regression (SR) can be effective in providing survival models that are both accurate and interpretable.
We propose an adaptation to SuR of a fully-fledged, multi-objective multi-expression SR algorithm based on genetic programming (GP)~\cite{gporiginalpaper,gptutorial1,gptutorial2,o2009riccardo}.
To the best of our knowledge, only a limited number of works exist on addressing SuR with SR (see \Cref{sec:related}).
The contributions of this paper are:
\begin{itemize}
\item We propose a GP-based search algorithm to adapt SR to SuR, in a multi-objective formulation with accuracy vs.~simplicity;
\item We propose procedures to obtain Pareto fronts from two traditional glass-box SuR methods in order to compare accuracy-simplicity trade-offs with SR;
\item We experimentally show that our SR approach can lead to SuR models with superior (respectively, similar) predictive performance compared to traditional glass-box (resp.,~black-box) models; while appearing promising in terms of interpretability.
\end{itemize}


\section{Background}
\label{sec:background}

In this section, we introduce foundations and review related work.

\subsection{Survival regression (SuR)}
\label{sec:sur}

Survival regression (SuR) involves the analysis and prediction of time to event data~\cite{miller2011survival}. 
We denote a survival dataset by $\mathcal{D} = \{ (\mathbf{x}_i, t_i, \delta_i) \}^n_i$.
Each row, indexed by $i$, represents an entity (e.g., patient); columns contain $d$ features $\mathbf{x}_i=( x_{1,i}, x_{2,i}, \dots, x_{d, i})^\intercal$ of the entity (e.g., age, weight, tumor stage), as well as a time $t_i$, and a \emph{censoring indicator} $\delta_i \in \{0,1\}$. 
The time $t_i$ refers to the onset of the adverse event (e.g., tumor progression or death) when $\delta_i=1$, while it refers to censoring (e.g., because the patient stopped the follow-up) when $\delta_i=0$.
Clearly, a complication of SuR over traditional regression is that censoring must be accounted for.
The scenario just described is referred to as \emph{right-censoring} and is perhaps the most common in survival applied to healthcare.
Regarding left-censoring and interval-censoring, which are not considered in this work, we refer to~\cite{leung1997censoring}.

To learn a predictive model from SuR data, let us start by considering the survival function $S$ and the hazard function $h$.
The former is:
\begin{equation}
    S(t) = Pr(T > t)
\end{equation}
and represents the probability of surviving (i.e., the adverse event has not happened) up to time $t$.
In turn, the hazard is:
\begin{equation}\label{eq:hazard}
    h(t) = -\frac{d \log S(t)}{dt}
\end{equation}
and represents the probability for an entity that has survived until $t$, that the event will happen at $t$~\cite{clark2003survival}.
Hereon we use $S(t, \mathbf{x})$ and $h(t, \mathbf{x})$ to denote that survival and hazard depend on the features.

We proceed by considering the traditional formulation in machine learning whereby the parameters $\bm\theta$ of the model that best explain the data must be found.
The likelihood for survival data is~\cite{lin2007breslow}:
\begin{equation}\label{eq:likelihood}
    L(\bm\theta) = \prod_i h(t_i, \mathbf{x_i} | \bm\theta)^{\delta_i} S(t_i, \mathbf{x_i} | \bm\theta).
\end{equation}
In other words, $\bm\theta$ must correctly describe the cases where the event (resp., censoring) happened at $t_i$, corresponding to $\delta_i=1$ (resp., $\delta_i = 0$), and thus contributing by the probability of surviving until exactly $t_i$, i.e., $h(t_i) S(t_i)$ (resp., surviving beyond $t_i$, i.e., $S(t_i)$).

To simplify the implementation and optimization of the hazard function in $\Cref{eq:likelihood}$, Sir David Cox famously proposed the \emph{proportional hazard assumption}~\cite{cox1972regression}, i.e., the ratio of hazards between two groups stays the same over time.
Under this assumption:
\begin{equation}\label{eq:cox}
    h(t, \mathbf{x}) = h_0(t)\exp(\bm\theta^\intercal \mathbf{x}),
\end{equation}
i.e., the hazard can be broken down into the \emph{baseline hazard} $h_0$ that depends only on $t$, and a proportional contribution given by exponentiation of the product between parameters $\bm\theta$ and features $\mathbf{x}$. 
\Cref{eq:cox} is called Cox proportional hazard model. 
In this model, $\bm\theta$ can be found by optimizing the \emph{partial likelihood} $L_p$~\cite{cox1975partial}:
\begin{equation}
    L_p(\bm\theta) = \prod_{i} \left[\frac{\exp(\bm\theta^\intercal \mathbf{x}_i )}{\sum_{j \in R(t_i)} \exp(\bm\theta^\intercal \mathbf{x}_j)}\right]^{\delta_i},
\end{equation}
where $R(t_i)$ is the \emph{risk set}, i.e., the set of entities still surviving at $t_i$.
Meanwhile, $h_0(t)$ can be realized by any non-negative function and can be optimized using methods such as the Breslow estimator, Efron estimator, or Kalbfleisch Prentice estimator~\cite{breslow1972discussion,lin2007breslow,efron1977efficiency,kalbfleisch1973marginal,kalbfleisch2002statistical}.

As Cox's model does not assume any specific form for the baseline hazard function, it is less restrictive than fully parametric models which can be misspecified and lead to biased predictions.
At the same time, the proportional hazard assumption can be incorrect, i.e., the ratio of hazards between two groups might change as time passes~\cite{cheung2021beware,kuitunen2021testing,hess1995graphical}.
Some recent machine learning-based proposals still rely and build on top of the proportional hazards assumption~\cite{ridgeway1999state,katzman2018deepsurv,nagpal2021deep,wilstrup2022combining}, while others attempt to drop it~\cite{ishwaran2008random,chen2013gradient,archetti2024fpboost}.



\subsection{Symbolic regression \& genetic programming}
\label{sec:sr}

Symbolic Regression (SR) is the problem of discovering mathematical expressions that best describe a given dataset~\cite{kronberger2024symbolic}.
Unlike traditional regression which considers parametric models, in SR a predefined model structure is not assumed, and both the structure and the parameters must be found.
Optimizing the parameters (or, as commonly called in symbolic regression, simply \emph{constants}) $\bm c$ can be achieved with traditional optimization methods, e.g., gradient-based when the structure is differentiable~\cite{harrison2023mini}.
For the structure, a set of primitive operations such as $+,-,\times,\div,\log,\sin,\dots$ must be chosen, and combined with both features $x_1, x_2, \dots$ and constants $c_1, c_2, \dots$ into a meaningful expression.

The advantage of SR is that SR models can be non-linear, thus fitting the data with high accuracy, while also potentially interpretable, e.g., when composed of a limited number of operations, features, and constants~\cite{virgolin2021modellearning,nadizar2024ananalysis}.
However, an important disadvantage is that the structure optimization aspect makes of SR an NP-hard problem~\cite{virgolin2022symbolic}.
While a variety of SR algorithms exist, including deep learning-based ones~\cite{landajuela2022unified,kim2020integration,zhang2023deep,d2022deep,kamienny2022end,vastl2024symformer}, those based on genetic programming (GP)~\cite{gporiginalpaper} often achieve state-of-the-art results~\cite{la2021contemporary}.
GP is an approach inspired by evolution, where a population of candidate programs (or, in this context, models) adapts by recombination and mutation of their atomic components, and selection of the fittest, over a number of generations.

\subsection{Related work}\label{sec:related}
A variety of different algorithms exist to deal with SuR.
Black-box non-linear algorithms include random survival forests~\cite{ishwaran2008random}, gradient boosting survival machines~\cite{chen2013gradient,archetti2024fpboost}, as well as deep learning-based methods~\cite{katzman2018deepsurv,nagpal2021deep}.
Glass-box methods include linear approaches such as (regularized) Cox's proportional hazard model~\cite{cox1972regression,simon2011regularization} and accelerated failure time models~\cite{kalbfleisch2002statistical,wei1992accelerated}; and non-linear approaches among which GAMs~\cite{Bai2017} and survival trees~\cite{leblanc1993survival,Bou-Hamad2011}.
Woodward et al.~propose a rule-based learning classifier system for survival~\cite{woodward2024survival}.

Regarding SR applications to SuR, the work by Wilstrup and Cave~\cite{wilstrup2022combining} is arguably the most similar to ours.
However, like in GAMs and differently from us, the authors use SR to discover only univariate non-linear functions.
Moreover, these functions are optimized independently from one another, and then set as basis functions for a Cox model.
We propose a multi-objective and multi-expression formulation where each non-linear function can take an arbitrary number of features, and is optimized simultaneously within the Cox model (see \Cref{sec:algorithm}).

Lastly, SR has been assessed to predict residual lifetime (or ``endurance'') of hardware, such as Flash devices~\cite{hogan2013mclnandflash}, turbofan engines~\cite{adams2019dataprognostics}, lithium-ion-cells~\cite{schofer2022machine}, and slewing bearings~\cite{ding2019asymbolic}.
Importantly, these works do not feature data with (right-)censoring, thus a traditional formulation of SR is taken, where only the error between the predicted and actual time needs to be considered.



\section{SR algorithm}
\label{sec:algorithm}

This section describes our multi-objective multi-expression GP-based adaptation of SR to SuR.

\subsection{Multi-expression representation}
\label{sec:multitree}
We adopt the proportional hazard assumption and seek to fit:
\begin{equation}\label{eq:our-cox}
\begin{split}
    h(t, \mathbf{x}) = h_0(t) \exp \left(\bm \theta^\intercal \bm f(\mathbf{x}) \right), \\
    \text{where \ \ } \bm \theta^\intercal \bm f(\mathbf{x}) = \sum_j \theta_j f_j(\mathbf{x}),
\end{split}
\end{equation}
i.e., we modify the Cox proportional hazard model to linearly combine  functions $f_1, f_2, \dots$ of the features instead of the features directly.
We represent each function $f_j$ as a mathematical expression composed of primitive operations, whose structure is optimized by GP.
The specific features used in an $f_j$ depend on its structure.
We refer to the number of distinct features used across expressions in the model as the model's dimensionality.

We set the population of GP to be composed of models, each following \Cref{eq:our-cox}.
We use GP's recombination and mutation operators to alter the structure and parameters of the expressions within the model, while we use coordinate descent to fit the parameters $\bm\theta$ that linearly combine the evolved functions~\cite{wright2015coordinate}.
We represent expressions with trees, encoding primitive operations, features, and constants with tree nodes~\cite{o2009riccardo}.

Our approach can be seen as a form of feature construction (similarly to, e.g., La Cava et al.~\cite{la2023flexible} for regression and Tran et al.~\cite{tran2019genetic} for classification), where $\bm f(\mathbf{x})$ are the constructed features, and the remaining terms in~\Cref{eq:our-cox} make the model for which these features are evolved.

\subsection{Multi-objective evolution}
We set GP to work in a multi-objective fashion, to discover models with trade-offs between accuracy and interpretability.
Specifically, we use the following objectives:
\begin{itemize}
    \item $\textit{obj}_1 (\uparrow)$: Concordance index for right-censored data based on inverse probability of censoring weights (\CI).
    In a nutshell, \CI assesses that the model's ability to predict survival order correctly, and is a well-established metric in SuR~\cite{harrell1996multivariable}.
    \item $\textit{obj}_2 (\downarrow)$: The number of \emph{dimensions} (i.e., distinct features) $x_1, x_2, \dots$ appearing in the model.
\end{itemize}
We set $\textit{obj}_2$ to the number dimensions rather than, e.g., the number of terms in the expressions as in~\cite{la2021contemporary,virgolin2021modellearning}, because:
(1) the number of terms can simply be constrained (see \Cref{sec:setting});
(2) interpreting a larger but lower-dimensional expression might be easier than interpreting a smaller but higher-dimensional expressions because one can reason by \emph{decomposition} of the contributions happening across dimensions~\cite{lipton2017doctor,lipton2018mythos};
(3) we find that reducing the number of dimensions anyway correlates with reducing the number of overall terms (see e.g.~\Cref{fig:multitreelengthlineplot}, \Cref{tab:gpformulae});
(4) this allows us to compare with survival trees, which are fundamentally different from expressions;
(5) in practical application such as in clinical trials, reducing the number of different patient features to be monitored can reduce costs and improve reliability.

Using the objectives above, we follow the Non-dominated Sorting Genetic Algorithm 2 (NSGA-2)~\cite{nsga2} to realize the multi-objective evolution.
We use duplicate penalization as a simple but effective way to contrast NSGA-2's tendency to over-duplicate small and hard to evolve expressions in GP~\cite{liu2022evolvability}.
To determine duplication, we compare the vectors of model predictions on the training set.

\section{Pareto fronts for other glass-box methods}
\label{sec:other-pareto}
We consider Cox's proportional hazard model with elastic net regularization (CX)~\cite{ebrahimi2022predictive} and survival trees (ST)~\cite{gordon1985tree} as glass-boxes for benchmarking.
We further consider survival adaptations of gradient boosting (GB)~\cite{chen2013gradient} and random forest (RF)~\cite{ishwaran2008random} as black-boxes.
All methods are implemented using the scikit-survival library~\cite{polsterl2020scikit}.

Black-box models are not interpretable and therefore for those we focus solely on \CI.
Conversely, since CX and ST models may include a different number of dimensions, we propose an approach to obtain Pareto fronts for each of them, enabling direct comparisons with the fronts obtained for SR.
This way, we can assess whether one approach is superior to another when more or less features are allowed.

For CX, we set the $L1$ ratio hyper-parameter, which balances between $L1$ and $L2$ regularization, to a fixed and typical value (of $0.5$~\cite{polsterl2020scikit}).
We then optimize CX varing the strength of regularization $\lambda$ among \num{1000} possible values, resulting in models with varying number of dimensions.
When multiple $\lambda$ values lead to same-dimensional models, we consider the model with median $\lambda$ value for the Pareto front.
For ST, we consider a range of maximal tree depths from $1$ to $25$: for each, we perform 5-fold grid-search to optimize the other hyper-parameters of the ST (see \Cref{sec:setting}).
After completing all \num{25} grid-search optimizations, the models are iterated in order of increasing maximum depth, calculating the number of dimensions; when multiple models are found with the same number of dimensions, the one with the smallest depth is picked for the Pareto front.

\begin{figure*}[!h]
    \hspace*{-0.5cm}\includegraphics[scale=0.41]{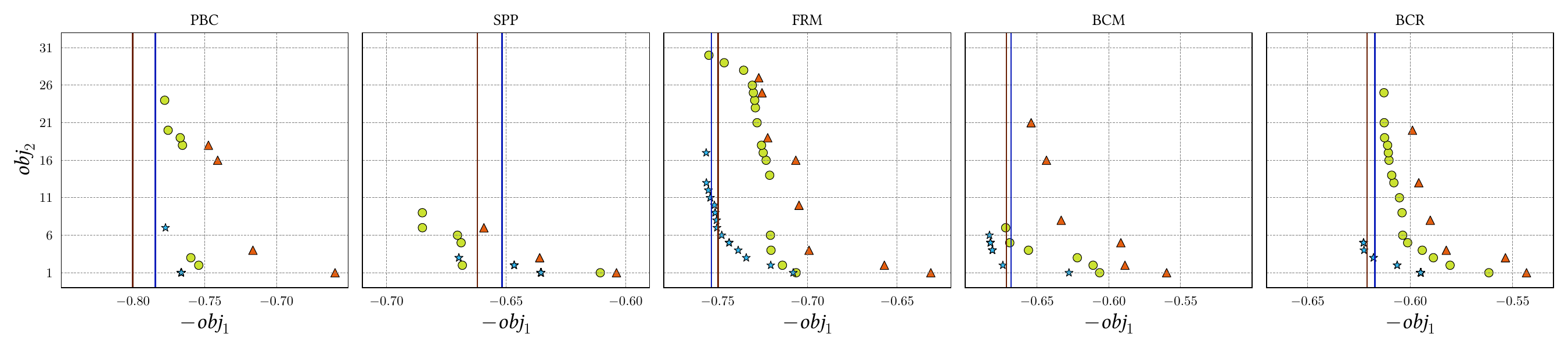}
    \\
    \hspace*{0.45cm}\includegraphics{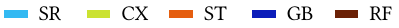}
    \caption{
    Pareto fronts with median test \HV for each dataset (normalized).
    We consider minimization of both objectives (low-left is best) for ease of interpretation.
    For black-box methods, which do not have fronts, the negated \CI is reported.
    }
    \label{fig:paretoallmethodsplot}
\end{figure*}

\section{Experiments}
\label{sec:exps}
In this section, we detail our experimental setup, including how specific aspects such as categorical features are handled.

\subsection{Data}\label{sec:data}
We consider five real-world datasets: PBC2 (\PBC) with \num{1945} observations and \num{19} features~\cite{therneau2000cox}, Support2 (\SPP) with \num{9105} observations and \num{46} features~\cite{support2}, Framingham (\FRM) with \num{11627} observations and \num{37} features~\cite{tsao2015framingham}, Breast Cancer Metabric (\BCM) with \num{2509} observations and \num{31} features~\cite{curtis2012genomic,pereira2016somatic}, and Breast Cancer Metabric Relapse (\BCR) with \num{2509} observations and \num{31} features~\cite{rueda2019dynamics}.

For each dataset, we consider two scenarios that can impact model accuracy and interpretability: using or not using $z$-score normalization (also called standardization)~\cite{harrison2023mini}.
On the one hand, standardizing makes features similarly-scaled, enabling e.g.~easy interpretation of parameter comparisons in CX (e.g., $2\times\texttt{age}$ vs.~$1\times\texttt{weight}$ signifies age contributes twice as much than weight).
On the other hand, when looking at the CX model as a whole, or at the decision nodes of an ST model, standardization might harm interpretation as one must consider that the parameters are relative to $\frac{x - \mu(x)}{\sigma(x)}$ instead of $x$.

SuR data often includes categorical features.
We handle the encoding of categorical features as follows:
if only two categories are present, we convert the categories to 0 (false) or 1 (true); 
else if categories are ordinal (e.g., \texttt{stage I}, \texttt{stage II}, etc.\ for feature \texttt{cancer stage}), convert the categories to integers starting from 0;
else, we use one-hot encoding.

\subsection{Hyper-parameter settings}
\label{sec:setting}
\subsubsection{SR algorithm}
For our GP-based SR algorithm, we set the population size $n_{\text{pop}}$ to \num{1000}, and run the evolution over \num{100} generations.
To promote interpretability, beyond the aformentioned $\textit{obj}_2$, we constrain the trees (which are used to represent expressions) to contain a maximum of \num{7} nodes.
Trees are initialized using the ramped half-and-half method~\cite{gporiginalpaper,gptutorial1,gptutorial2}.
We initialize the models to contain \num{1} to \num{4} trees (expressions), uniformly at random. 

We use $+,\ -,\ \times,\ \text{Square},\ \text{ProtectedLog},\ \text{AQ}$ as primitive operations\footnote{
    $\text{AQ}(a, b) = \frac{a}{\sqrt{b^2 + 1}}$,
    $\text{ProtectedLog}(a) = \log(|a| + 10^{-9})$, to prevent numerical errors.
}.
Additionally, the features of the dataset $x_1, x_2, \dots$ (normalized or encoded as per \Cref{sec:data}) and ephemeral random constants~\cite{o2009riccardo} uniformly sampled within $[-5, 5]$ are used as tree nodes to represent variables and constants in the expressions.
We treat features containing $0$-$1$ values specially: we mimic linear models by enforcing a couple of these features with a coefficient, using tree nodes that implement $x_i \times c$, with $c \in \mathbb{R}$ a constant whose value is sampled when the node is initialized, as in ephemeral random constants.

We use a tournament size of \num{4} for the selecting parents as per NSGA-2. 
To alter the structure of offspring models we use a cocktail of recombination and mutation operators.
These are
\emph{expression addition/deletion}: a randomly-initialized tree is added or a random tree is removed from the existing ones (each with probability of $0.05$); \emph{expression crossover}: a random tree is discarded and a random tree from a random donor model is cloned and added (prob.~$0.1$);
\emph{sub-tree crossover}: like the previous, however at the level of sub-trees (prob.~$0.1$);
\emph{node-level crossover}: like the previous, however at the level of nodes that are compatible, i.e., share the same number of inputs (prob.~$0.25$);
\emph{sub-tree mutation}: like sub-tree crossover, but the replacing sub-tree is initialized at random instead of cloned from a donor (prob.~$0.25$); \emph{node-level mutation}: like node-level crossover, but the replacing node is random instead of cloned from a donor (prob.~$0.25$).
The order of application of these operators is randomized, and only one is applied.
Afterwards, we stochastically apply to 90\% of the offspring constant mutation, where a constant node has probability of $0.5$ being altered, using a temperature\footnote{
    The new constant value is computed as $c + t|c| \sim N(0,1)$, where $t$ is the temperature.
} of $0.1$, which is relatively easy to implement and was found to be competitive with gradient-based optimization ~\cite{harrison2023mini}.
After structural and constant changes, the parameters $\bm \theta$ of \Cref{eq:our-cox} are fitted with coordinate descent~\cite{polsterl2020scikit}.
In particular we use the same implementation and settings of CX (see \Cref{sec:other-pareto}), with $\lambda$ set to a small value ($10^{-6}$).

We note that we resort to fixing the hyper-parameters as described, instead of using hyper-parameter tuning, because our algorithm takes ca.~1 hour per evolution (implemented in Python, run on Intel(R) Xeon(R) W-2295 CPU and 64 GB RAM).

\subsubsection{Competing algorithms}
We refer back to \Cref{sec:other-pareto} for the settings of CX.
For ST, GB, and RF, the models are trained using a grid-search approach with cross-validation (on the training set), optimizing \CI over \num{5} folds. \Cref{tab:grids} reports the hyper-parameter options we adopt.

\begin{table}[ht!]
\centering
\caption{Hyper-parameter grids for Survival Tree and survival adaptations of Gradient Boosting and Random Forest.}
\label{tab:grids}
\small
\begin{adjustbox}{max width=\linewidth}
\begin{tabular}{@{}ll@{}}
\toprule
\textbf{Model}                        & \textbf{Hyper-parameter Grid} \\ 
\midrule
\textbf{Survival Tree (ST)}                & 
\begin{minipage}[t]{0.7\linewidth}
\begin{itemize}
    \item \texttt{min\_samples\_split: [2, 5, 8]}
    \item \texttt{min\_samples\_leaf: [1, 4]}
    \item \texttt{max\_features: [0.5, 1.0]}
    \item \texttt{splitter: ['best', 'random']}
\end{itemize}
\end{minipage} \\
\midrule
\textbf{Gradient Boosting (GB)}            & 
\begin{minipage}[t]{0.7\linewidth}
\begin{itemize}
    \item \texttt{max\_depth: [3, 6, 9]}
    \item \texttt{loss: ['coxph', 'ipcwls']}
    \item \texttt{learning\_rate: [0.1, 0.01]}
    \item \texttt{n\_estimators: [50, 250]}
    \item \texttt{min\_samples\_split: [2, 5, 8]}
    \item \texttt{min\_samples\_leaf: [1, 4]}
\end{itemize}
\end{minipage} \\
\midrule
\textbf{Random Forest (RF)}                & 
\begin{minipage}[t]{0.7\linewidth}
\begin{itemize}
    \item \texttt{max\_depth: [3, 6, 9]}
    \item \texttt{n\_estimators: [50, 250]}
    \item \texttt{min\_samples\_split: [2, 5, 8]}
    \item \texttt{min\_samples\_leaf: [1, 4]}
\end{itemize}
\end{minipage} \\
\bottomrule
\end{tabular}
\end{adjustbox}

\end{table}

\begin{figure}[!htbp]
    \hspace*{-0.5cm}\includegraphics[scale=0.45]{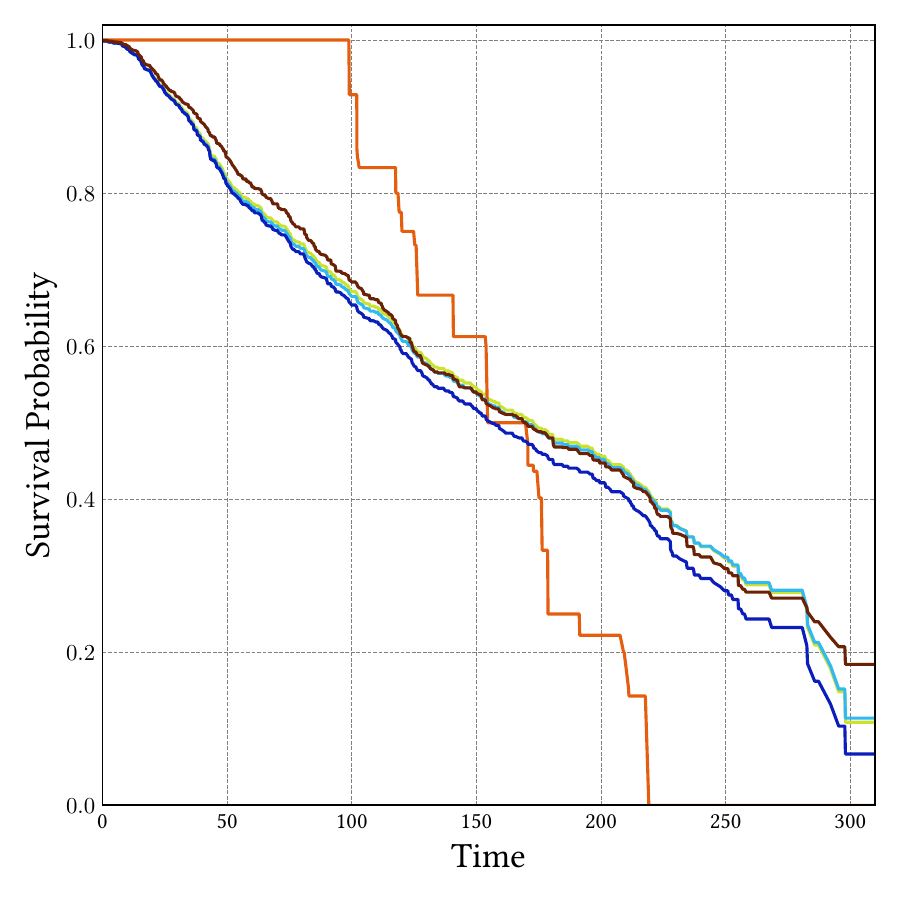}
    \\
    \hspace*{0.72cm}\includegraphics{legendamethods.pdf}
    \caption{
    Median probability of survival (across observations i.e.~patients) of a random repetition on (normalized) \BCM test set.
    For glass-box methods, we take the highest dimensional model from the Pareto front.
    }
    \label{fig:survfuncplot}
\end{figure}

\subsection{Assessment}
For each combination of method, dataset, and hyper-parameters, \num{50} independent repetitions are performed. Specifically, each dataset is split into \num{50} random train-test partitions using a \num{7}:\num{3} ratio.

From this point onward, we use $k$ to denote the number of dimensions of a model from a given Pareto front.
We will focus on $k \in [3..7]$ following Miller's Law on the number of objects that can be considered by humans~\cite{miller1956magical}.
We use the notation ``$\text{max}$'' when considering the model in the Pareto front with the highest number of dimensions.

To evaluate the quality of a Pareto front, we employ the hyper-volume (\HV) as it measures the size of the space covered by the models in the front in terms of both objectives~\cite{zitzler2004indicator}.
A higher \HV indicates a better overall performance.
To focus on differences regarding simpler models in the fronts, some results are reported for models with exactly or up $k$ dimensions from the Pareto.

To assess statistical significance, we use  Kruskal-Wallis~\cite{kruskal1952use} ($\alpha = 0.05$) across methods and datasets, followed by pairwise Wilcoxon-Mann-Whitney tests~\cite{mann1947test} ($\alpha = 0.05$) with Holm-Bonferroni correction~\cite{holm1979bonferroni} to compare pairs of methods on a same dataset.
We use a black asterisk (${\scalebox{0.90}{\textbf{\color{black}*}}}$) to mark methods that outperform at least one other method in the group according to the pairwise test, and a blue asterisk (${\scalebox{0.90}{\textbf{\color{blue}*}}}$) for methods that outperform all other methods.

\section{Results}
\label{sec:results}

Our results are presented by first considering performance, i.e., \CI, number of dimensions, and \HV, and then evaluating the readability and interpretability of the models found with our SR algorithm.

\begin{table*}[h]
    \centering
    \renewcommand{\arraystretch}{0.8}
    \caption{Median values (across the repetitions) of \HV from the Pareto front computed on the test set after the end of the optimization for the glass-box methods. 
    }
    \label{tab:hvtable}
    \scalebox{1.0}{
    \begin{tabular}{
    c
    c
    S[table-format=1.3]
    S[table-format=1.3]
    S[table-format=1.3]
    S[table-format=1.3]
    S[table-format=1.3]
    S[table-format=1.3]
    S[table-format=1.3]
    S[table-format=1.3]
    S[table-format=1.3]
    S[table-format=1.3]
    }
         \toprule
 &  & \multicolumn{5}{c}{Normalization} & \multicolumn{5}{c}{No Normalization} \\
 \cmidrule(lr){3-7} \cmidrule(lr){8-12}
 Up to $k$ & & {\PBC} & {\SPP} & {\FRM} & {\BCM} & {\BCR} & {\PBC} & {\SPP} & {\FRM} & {\BCM} & {\BCR} \\
         \midrule

\multirow{3}{*}{3} & ST & 68.324 & 63.015 & 67.027 & 60.702 & 56.234 & 68.138{$^{\scalebox{0.90}{\textbf{\color{black}*}}}$} & 63.015{$^{\scalebox{0.90}{\textbf{\color{black}*}}}$} & 67.027 & 60.702 & 56.234 \\ 
  & CX & 73.609{$^{\scalebox{0.90}{\textbf{\color{black}*}}}$} & 63.459 & 70.828{$^{\scalebox{0.90}{\textbf{\color{black}*}}}$} & 62.742{$^{\scalebox{0.90}{\textbf{\color{black}*}}}$} & 59.117{$^{\scalebox{0.90}{\textbf{\color{black}*}}}$} & 59.728 & 50.05 & 70.634{$^{\scalebox{0.90}{\textbf{\color{black}*}}}$} & 65.091{$^{\scalebox{0.90}{\textbf{\color{black}*}}}$} & 59.145{$^{\scalebox{0.90}{\textbf{\color{black}*}}}$} \\ 
  & SR & \bfseries 75.7{$^{\scalebox{0.90}{\textbf{\color{blue}*}}}$} & \bfseries 65.475{$^{\scalebox{0.90}{\textbf{\color{blue}*}}}$} & \bfseries 72.874{$^{\scalebox{0.90}{\textbf{\color{blue}*}}}$} & \bfseries 66.691{$^{\scalebox{0.90}{\textbf{\color{blue}*}}}$} & \bfseries 60.924{$^{\scalebox{0.90}{\textbf{\color{blue}*}}}$} & \bfseries 75.526{$^{\scalebox{0.90}{\textbf{\color{blue}*}}}$} & \bfseries 65.295{$^{\scalebox{0.90}{\textbf{\color{blue}*}}}$} & \bfseries 72.854{$^{\scalebox{0.90}{\textbf{\color{blue}*}}}$} & \bfseries 66.439{$^{\scalebox{0.90}{\textbf{\color{blue}*}}}$} & \bfseries 60.925{$^{\scalebox{0.90}{\textbf{\color{blue}*}}}$} \\ 
 \midrule 
\multirow{3}{*}{5} & ST & 70.948 & 63.066 & 68.15 & 62.501 & 57.067 & 70.959{$^{\scalebox{0.90}{\textbf{\color{black}*}}}$} & 63.066{$^{\scalebox{0.90}{\textbf{\color{black}*}}}$} & 68.15 & 62.501 & 57.067 \\ 
  & CX & 73.755{$^{\scalebox{0.90}{\textbf{\color{black}*}}}$} & 64.81{$^{\scalebox{0.90}{\textbf{\color{black}*}}}$} & 71.319{$^{\scalebox{0.90}{\textbf{\color{black}*}}}$} & 65.613{$^{\scalebox{0.90}{\textbf{\color{black}*}}}$} & 59.771{$^{\scalebox{0.90}{\textbf{\color{black}*}}}$} & 62.564 & 50.05 & 72.319{$^{\scalebox{0.90}{\textbf{\color{black}*}}}$} & 65.312{$^{\scalebox{0.90}{\textbf{\color{black}*}}}$} & 59.547{$^{\scalebox{0.90}{\textbf{\color{black}*}}}$} \\ 
  & SR & \bfseries 76.419{$^{\scalebox{0.90}{\textbf{\color{blue}*}}}$} & \bfseries 65.612{$^{\scalebox{0.90}{\textbf{\color{blue}*}}}$} & \bfseries 73.713{$^{\scalebox{0.90}{\textbf{\color{blue}*}}}$} & \bfseries 67.092{$^{\scalebox{0.90}{\textbf{\color{blue}*}}}$} & \bfseries 61.455{$^{\scalebox{0.90}{\textbf{\color{blue}*}}}$} & \bfseries 76.154{$^{\scalebox{0.90}{\textbf{\color{blue}*}}}$} & \bfseries 65.684{$^{\scalebox{0.90}{\textbf{\color{blue}*}}}$} & \bfseries 73.726{$^{\scalebox{0.90}{\textbf{\color{blue}*}}}$} & \bfseries 66.718{$^{\scalebox{0.90}{\textbf{\color{blue}*}}}$} & \bfseries 61.637{$^{\scalebox{0.90}{\textbf{\color{blue}*}}}$} \\ 
 \midrule 
\multirow{3}{*}{7} & ST & 71.494 & 63.806 & 68.401 & 62.98 & 57.595 & 71.418{$^{\scalebox{0.90}{\textbf{\color{black}*}}}$} & 63.806{$^{\scalebox{0.90}{\textbf{\color{black}*}}}$} & 68.401 & 62.98 & 57.634 \\ 
  & CX & 73.755{$^{\scalebox{0.90}{\textbf{\color{black}*}}}$} & \bfseries 66.64{$^{\scalebox{0.90}{\textbf{\color{black}*}}}$} & 71.369{$^{\scalebox{0.90}{\textbf{\color{black}*}}}$} & 65.892{$^{\scalebox{0.90}{\textbf{\color{black}*}}}$} & 59.833{$^{\scalebox{0.90}{\textbf{\color{black}*}}}$} & 62.564 & 50.05 & 72.348{$^{\scalebox{0.90}{\textbf{\color{black}*}}}$} & 65.312{$^{\scalebox{0.90}{\textbf{\color{black}*}}}$} & 59.547{$^{\scalebox{0.90}{\textbf{\color{black}*}}}$} \\ 
  & SR & \bfseries 76.565{$^{\scalebox{0.90}{\textbf{\color{blue}*}}}$} & 66.141{$^{\scalebox{0.90}{\textbf{\color{black}*}}}$} & \bfseries 74.217{$^{\scalebox{0.90}{\textbf{\color{blue}*}}}$} & \bfseries 67.228{$^{\scalebox{0.90}{\textbf{\color{blue}*}}}$} & \bfseries 61.455{$^{\scalebox{0.90}{\textbf{\color{blue}*}}}$} & \bfseries 76.297{$^{\scalebox{0.90}{\textbf{\color{blue}*}}}$} & \bfseries 66.012{$^{\scalebox{0.90}{\textbf{\color{blue}*}}}$} & \bfseries 74.164{$^{\scalebox{0.90}{\textbf{\color{blue}*}}}$} & \bfseries 67.026{$^{\scalebox{0.90}{\textbf{\color{blue}*}}}$} & \bfseries 61.693{$^{\scalebox{0.90}{\textbf{\color{blue}*}}}$} \\ 
 \midrule 
\multirow{3}{*}{\text{max}} & ST & 73.338 & 65.162 & 71.328 & 64.02 & 58.969 & 73.362{$^{\scalebox{0.90}{\textbf{\color{black}*}}}$} & 65.162{$^{\scalebox{0.90}{\textbf{\color{black}*}}}$} & 71.338 & 64.064 & 59.025 \\ 
  & CX & 75.904{$^{\scalebox{0.90}{\textbf{\color{black}*}}}$} & \bfseries 67.696{$^{\scalebox{0.90}{\textbf{\color{blue}*}}}$} & 73.809{$^{\scalebox{0.90}{\textbf{\color{black}*}}}$} & 66.308{$^{\scalebox{0.90}{\textbf{\color{black}*}}}$} & 60.401{$^{\scalebox{0.90}{\textbf{\color{black}*}}}$} & 62.564 & 50.05 & 72.348{$^{\scalebox{0.90}{\textbf{\color{black}*}}}$} & 65.496{$^{\scalebox{0.90}{\textbf{\color{black}*}}}$} & 60.132{$^{\scalebox{0.90}{\textbf{\color{black}*}}}$} \\ 
  & SR & \bfseries 76.865{$^{\scalebox{0.90}{\textbf{\color{black}*}}}$} & 66.37{$^{\scalebox{0.90}{\textbf{\color{black}*}}}$} & \bfseries 74.72{$^{\scalebox{0.90}{\textbf{\color{blue}*}}}$} & \bfseries 67.551{$^{\scalebox{0.90}{\textbf{\color{blue}*}}}$} & \bfseries 61.549{$^{\scalebox{0.90}{\textbf{\color{blue}*}}}$} & \bfseries 76.677{$^{\scalebox{0.90}{\textbf{\color{blue}*}}}$} & \bfseries 66.389{$^{\scalebox{0.90}{\textbf{\color{blue}*}}}$} & \bfseries 74.545{$^{\scalebox{0.90}{\textbf{\color{blue}*}}}$} & \bfseries 67.242{$^{\scalebox{0.90}{\textbf{\color{blue}*}}}$} & \bfseries 61.781{$^{\scalebox{0.90}{\textbf{\color{blue}*}}}$} \\ 
       \bottomrule
    \end{tabular}
    }
\end{table*}

\begin{table*}[h]
    \centering

    \renewcommand{\arraystretch}{0.8}
    \caption{Median values (across the repetitions) of \CI from the Pareto front computed on the test set after the end of the optimization for the glass-box methods.
    The trail (-) represents cases where no models were found with exactly $k$ dimensions.}
    \label{tab:citable}
    \begin{tabular}{
    c
    c
    S[table-format=1.3]
    S[table-format=1.3]
    S[table-format=1.3]
    S[table-format=1.3]
    S[table-format=1.3]
    S[table-format=1.3]
    S[table-format=1.3]
    S[table-format=1.3]
    S[table-format=1.3]
    S[table-format=1.3]
    }
         \toprule
 &  & \multicolumn{5}{c}{Normalization} & \multicolumn{5}{c}{No Normalization} \\
 \cmidrule(lr){3-7} \cmidrule(lr){8-12}
 $k$ & & {\PBC} & {\SPP} & {\FRM} & {\BCM} & {\BCR} & {\PBC} & {\SPP} & {\FRM} & {\BCM} & {\BCR} \\
         \midrule

\multirow{3}{*}{3} & ST & 0.676 & 0.635 & 0.669{$^{\scalebox{0.90}{\textbf{\color{black}*}}}$} & 0.616 & 0.568 & 0.676{$^{\scalebox{0.90}{\textbf{\color{black}*}}}$} & 0.635{$^{\scalebox{0.90}{\textbf{\color{black}*}}}$} & 0.669 & 0.616 & 0.568 \\ 
  & CX & 0.732{$^{\scalebox{0.90}{\textbf{\color{black}*}}}$} & 0.646 & {-} & 0.634{$^{\scalebox{0.90}{\textbf{\color{black}*}}}$} & 0.598{$^{\scalebox{0.90}{\textbf{\color{black}*}}}$} & 0.64 & {-} & 0.715{$^{\scalebox{0.90}{\textbf{\color{black}*}}}$} & 0.658{$^{\scalebox{0.90}{\textbf{\color{black}*}}}$} & 0.597{$^{\scalebox{0.90}{\textbf{\color{black}*}}}$} \\ 
  & SR & \bfseries 0.758{$^{\scalebox{0.90}{\textbf{\color{blue}*}}}$} & \bfseries 0.652{$^{\scalebox{0.90}{\textbf{\color{blue}*}}}$} & \bfseries 0.736{$^{\scalebox{0.90}{\textbf{\color{blue}*}}}$} & \bfseries 0.668{$^{\scalebox{0.90}{\textbf{\color{blue}*}}}$} & \bfseries 0.615{$^{\scalebox{0.90}{\textbf{\color{blue}*}}}$} & \bfseries 0.756{$^{\scalebox{0.90}{\textbf{\color{blue}*}}}$} & \bfseries 0.648{$^{\scalebox{0.90}{\textbf{\color{blue}*}}}$} & \bfseries 0.736{$^{\scalebox{0.90}{\textbf{\color{blue}*}}}$} & \bfseries 0.669{$^{\scalebox{0.90}{\textbf{\color{blue}*}}}$} & \bfseries 0.615{$^{\scalebox{0.90}{\textbf{\color{blue}*}}}$} \\ 
 \midrule 
\multirow{3}{*}{5} & ST & 0.719 & 0.656 & 0.693{$^{\scalebox{0.90}{\textbf{\color{black}*}}}$} & 0.639 & 0.582 & 0.719{$^{\scalebox{0.90}{\textbf{\color{black}*}}}$} & \bfseries 0.656{$^{\scalebox{0.90}{\textbf{\color{black}*}}}$} & 0.693 & 0.639 & 0.582 \\ 
  & CX & 0.732{$^{\scalebox{0.90}{\textbf{\color{black}*}}}$} & \bfseries 0.657 & {-} & 0.662{$^{\scalebox{0.90}{\textbf{\color{black}*}}}$} & 0.604{$^{\scalebox{0.90}{\textbf{\color{black}*}}}$} & {-} & {-} & 0.733{$^{\scalebox{0.90}{\textbf{\color{black}*}}}$} & \bfseries 0.67{$^{\scalebox{0.90}{\textbf{\color{black}*}}}$} & 0.6{$^{\scalebox{0.90}{\textbf{\color{black}*}}}$} \\ 
  & SR & \bfseries 0.763{$^{\scalebox{0.90}{\textbf{\color{blue}*}}}$} & 0.652 & \bfseries 0.745{$^{\scalebox{0.90}{\textbf{\color{blue}*}}}$} & \bfseries 0.669{$^{\scalebox{0.90}{\textbf{\color{black}*}}}$} & \bfseries 0.615{$^{\scalebox{0.90}{\textbf{\color{blue}*}}}$} & \bfseries 0.765{$^{\scalebox{0.90}{\textbf{\color{blue}*}}}$} & 0.655{$^{\scalebox{0.90}{\textbf{\color{black}*}}}$} & \bfseries 0.745{$^{\scalebox{0.90}{\textbf{\color{blue}*}}}$} & 0.667{$^{\scalebox{0.90}{\textbf{\color{black}*}}}$} & \bfseries 0.614{$^{\scalebox{0.90}{\textbf{\color{blue}*}}}$} \\ 
 \midrule 
\multirow{3}{*}{7} & ST & 0.727 & 0.646 & 0.704{$^{\scalebox{0.90}{\textbf{\color{black}*}}}$} & 0.637 & 0.583 & 0.727{$^{\scalebox{0.90}{\textbf{\color{black}*}}}$} & 0.646{$^{\scalebox{0.90}{\textbf{\color{black}*}}}$} & 0.704{$^{\scalebox{0.90}{\textbf{\color{black}*}}}$} & 0.638 & 0.584 \\ 
  & CX & 0.74 & \bfseries 0.674{$^{\scalebox{0.90}{\textbf{\color{blue}*}}}$} & {-} & 0.662{$^{\scalebox{0.90}{\textbf{\color{black}*}}}$} & 0.604{$^{\scalebox{0.90}{\textbf{\color{black}*}}}$} & {-} & {-} & {-} & \bfseries 0.677{$^{\scalebox{0.90}{\textbf{\color{black}*}}}$} & 0.599{$^{\scalebox{0.90}{\textbf{\color{black}*}}}$} \\ 
  & SR & \bfseries 0.763{$^{\scalebox{0.90}{\textbf{\color{blue}*}}}$} & 0.656{$^{\scalebox{0.90}{\textbf{\color{black}*}}}$} & \bfseries 0.751{$^{\scalebox{0.90}{\textbf{\color{blue}*}}}$} & \bfseries 0.671{$^{\scalebox{0.90}{\textbf{\color{black}*}}}$} & \bfseries 0.613{$^{\scalebox{0.90}{\textbf{\color{blue}*}}}$} & \bfseries 0.764{$^{\scalebox{0.90}{\textbf{\color{blue}*}}}$} & \bfseries 0.659{$^{\scalebox{0.90}{\textbf{\color{blue}*}}}$} & \bfseries 0.75{$^{\scalebox{0.90}{\textbf{\color{blue}*}}}$} & 0.671{$^{\scalebox{0.90}{\textbf{\color{black}*}}}$} & \bfseries 0.613{$^{\scalebox{0.90}{\textbf{\color{blue}*}}}$} \\ 
 \midrule 
\multirow{3}{*}{\text{max}} & ST & 0.704 & 0.604 & 0.705 & 0.605 & 0.573 & 0.704{$^{\scalebox{0.90}{\textbf{\color{black}*}}}$} & 0.604{$^{\scalebox{0.90}{\textbf{\color{black}*}}}$} & 0.705 & 0.601 & 0.574 \\ 
  & CX & 0.768{$^{\scalebox{0.90}{\textbf{\color{black}*}}}$} & \bfseries 0.666{$^{\scalebox{0.90}{\textbf{\color{blue}*}}}$} & 0.754{$^{\scalebox{0.90}{\textbf{\color{black}*}}}$} & \bfseries 0.669{$^{\scalebox{0.90}{\textbf{\color{black}*}}}$} & 0.602{$^{\scalebox{0.90}{\textbf{\color{black}*}}}$} & 0.626 & 0.506 & 0.733{$^{\scalebox{0.90}{\textbf{\color{black}*}}}$} & 0.66{$^{\scalebox{0.90}{\textbf{\color{black}*}}}$} & 0.602{$^{\scalebox{0.90}{\textbf{\color{black}*}}}$} \\ 
  & SR & \bfseries 0.77{$^{\scalebox{0.90}{\textbf{\color{black}*}}}$} & 0.649{$^{\scalebox{0.90}{\textbf{\color{black}*}}}$} & \bfseries 0.756{$^{\scalebox{0.90}{\textbf{\color{black}*}}}$} & 0.664{$^{\scalebox{0.90}{\textbf{\color{black}*}}}$} & \bfseries 0.608{$^{\scalebox{0.90}{\textbf{\color{blue}*}}}$} & \bfseries 0.765{$^{\scalebox{0.90}{\textbf{\color{blue}*}}}$} & \bfseries 0.655{$^{\scalebox{0.90}{\textbf{\color{blue}*}}}$} & \bfseries 0.753{$^{\scalebox{0.90}{\textbf{\color{blue}*}}}$} & \bfseries 0.668{$^{\scalebox{0.90}{\textbf{\color{black}*}}}$} & \bfseries 0.611{$^{\scalebox{0.90}{\textbf{\color{blue}*}}}$} \\

         \bottomrule
    \end{tabular}
\end{table*}

\begin{table}[h]
    \centering
    \renewcommand{\arraystretch}{0.8}
    \caption{Median values (across the repetitions) of \CI on the test set for GB, RF, and SR.
    Black-box models employ all the features, while the SR model is the one with the maximum number of dimensions from the Pareto front.}
    \label{tab:citableblackbox}
    \begin{tabular}{
    c
    S[table-format=1.3]
    S[table-format=1.3]
    S[table-format=1.3]
    S[table-format=1.3]
    S[table-format=1.3]
    }
         \toprule
   & {\PBC} & {\SPP} & {\FRM} & {\BCM} & {\BCR} \\
         \midrule

 GB & 0.784{$^{\scalebox{0.90}{\textbf{\color{black}*}}}$} & 0.652 & 0.754{$^{\scalebox{0.90}{\textbf{\color{black}*}}}$} & 0.668 & 0.617{$^{\scalebox{0.90}{\textbf{\color{black}*}}}$}  \\ 
  RF & \bfseries 0.8{$^{\scalebox{0.90}{\textbf{\color{blue}*}}}$} & \bfseries 0.662 & 0.75 & \bfseries 0.671 & \bfseries 0.621{$^{\scalebox{0.90}{\textbf{\color{black}*}}}$}  \\ 
  SR & 0.77 & 0.649 & \bfseries 0.756{$^{\scalebox{0.90}{\textbf{\color{black}*}}}$} & 0.664 & 0.608  \\

         \bottomrule
    \end{tabular}
\end{table}

\begin{figure*}[!h]
    \centering
    \hspace*{0cm}\includegraphics[scale=0.7]{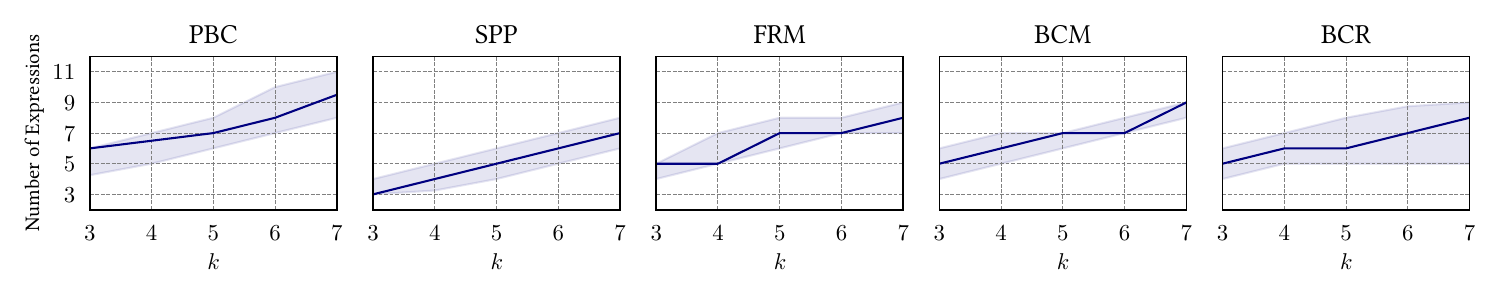}
    \caption{
    Median number of expressions (shaded area represents inter-quartile range) in the discovered SR models at different number of dimensions $k$.
    }
    \label{fig:multitreelengthlineplot}
\end{figure*}

\begin{table*}[h]
    \centering
    \renewcommand{\arraystretch}{1.0}
    \caption{Examples of the $\bm\theta^\intercal \bm f(\mathbf{x})$ obtained with SR for different number of dimensions $k$ (no normalization).
    }
    \label{tab:gpformulae}
    \begin{tabular}{p{0.025\textwidth} p{0.001\textwidth} S[table-format=1.3, table-column-width=0.03\textwidth]
        S[table-format=1.3, table-column-width=0.03\textwidth] >{\tiny}l}
         \toprule
 & & \multicolumn{2}{c}{\CI} & \\
         \cmidrule(lr){3-4}
     & $k$ & {Train} & {Test} & {} \\
         \midrule

\multirow{3}{*}{\PBC} & 3 & 0.795 & 0.715 & $- 0.074 x_{18}^{2} \log{\left(\left|{x_{1}}\right| \right)} - \frac{0.301 x_{18}^{2}}{x_{24}^{2} + 0.295} + 0.714 \log{\left(\left|{x_{1}}\right| \right)} + 0.174 \log{\left(\left|{\log{\left(\left|{x_{1}}\right| \right)}}\right| \right)} - 0.303$ \\ 
  & 5 & 0.81 & 0.741 & $- 1.082 x_{18}^{4} + \frac{0.185 x_{3}}{\sqrt{x_{10}^{2} + 0.051}} - 1.21 \log{\left(\left|{x_{1}}\right| \right)} \log{\left(\left|{x_{3}}\right| \right)} + 2.386 \log{\left(\left|{x_{1}}\right| \right)} + 0.445 \log{\left(\left|{x_{3}}\right| \right)}^{2} - 5.798 \log{\left(\left|{x_{3}}\right| \right)} + 0.436 \log{\left(\left|{x_{4}}\right| \right)} + 0.167 \log{\left(\left|{\log{\left(\left|{x_{1}}\right| \right)}}\right| \right)} - 2.881$ \\ 
  & 7 & 0.821 & 0.743 & $- \frac{1.18 x_{18}^{2}}{0.752 x_{24}^{2} + 1.0} + \frac{0.139 x_{3}}{\sqrt{x_{10}^{2} + 0.058}} + 0.572 \log{\left(\left|{x_{1}}\right| \right)} - 0.022 \log{\left(\left|{x_{22}}\right| \right)} - 4.067 \log{\left(\left|{x_{3}}\right| \right)} + 0.478 \log{\left(\left|{x_{4}}\right| \right)} + 0.108 \log{\left(\left|{\log{\left(\left|{x_{1}}\right| \right)}}\right| \right)} - 1.123$ \\ 
\midrule 
\multirow{3}{*}{\SPP} & 3 & 0.715 & 0.647 & $0.176 \log{\left(\left|{x_{1}}\right| \right)} + 0.06 \log{\left(\left|{3.428 x_{27} + 2.681 x_{41}}\right| \right)} - 0.053 \log{\left(\left|{\log{\left(\left|{x_{27}}\right| \right)} + 0.268}\right| \right)} - 1.195$ \\ 
  & 5 & 0.731 & 0.657 & $0.024 \left(x_{17} - 0.389 x_{25}\right)^{2} + 0.175 \log{\left(\left|{x_{1}}\right| \right)} - 0.059 \log{\left(\left|{x_{17}}\right| \right)} + 0.059 \log{\left(\left|{x_{17} + 1.862 x_{25}}\right| \right)} + 0.064 \log{\left(\left|{1.789 x_{27} + 4.592 x_{41}}\right| \right)} - 0.032 \log{\left(\left|{\log{\left(\left|{x_{27}}\right| \right)} + 0.378}\right| \right)} - 0.836$ \\ 
  & 7 & 0.739 & 0.647 & $0.188 x_{17} - 0.786 x_{24}^{2} + 0.11 x_{42}^{2} - 0.46 x_{42} - 0.054 \left(0.234 x_{17} + x_{42}\right)^{2} + 0.146 \log{\left(\left|{x_{1}}\right| \right)} + 0.058 \log{\left(\left|{1.875 x_{27} + 3.884 x_{41}}\right| \right)} + 0.138 \log{\left(\left|{\log{\left(\left|{x_{28}}\right| \right)} + 1.479}\right| \right)} - 0.779$ \\ 
\midrule 
\multirow{3}{*}{\FRM} & 3 & 0.737 & 0.734 & $- 0.745 x_{9}^{2} + 4.104 \log{\left(\left|{x_{1}}\right| \right)} + 46.549$ \\ 
  & 5 & 0.746 & 0.74 & $0.081 x_{1} - 0.431 x_{10} + 0.975 x_{13} - 0.589 x_{9}^{2} - 0.025 \log{\left(\left|{x_{28}}\right| \right)} + 6.771$ \\ 
  & 7 & 0.754 & 0.747 & $0.078 x_{1} + 0.426 x_{13} + 0.713 x_{20} x_{9}^{2} + 0.017 x_{28}^{2} - 0.021 \log{\left(\left|{x_{10}}\right| \right)} + 0.012 \log{\left(\left|{x_{13}}\right| \right)} - 0.041 \log{\left(\left|{x_{20}}\right| \right)} - 0.012 \log{\left(\left|{x_{28}}\right| \right)} - 0.057 \log{\left(\left|{x_{9}}\right| \right)} + 5.648$ \\ 
\midrule 
\multirow{3}{*}{\BCM} & 3 & 0.663 & 0.701 & $0.177 x_{0} - 4.947 \log{\left(\left|{x_{0}^{2}}\right| \right)} + 1.407 \log{\left(\left|{x_{0} + 0.046}\right| \right)} + 2.303 \log{\left(\left|{x_{3} + x_{6}}\right| \right)} - 1.434 \log{\left(\left|{x_{6} - 2.725}\right| \right)} - 20.835$ \\ 
  & 5 & 0.678 & 0.691 & $0.18 x_{0} + 0.512 x_{32}^{2} + 0.265 x_{58}^{4} - 8.657 \log{\left(\left|{x_{0}}\right| \right)} + 2.583 \log{\left(\left|{x_{3} + x_{6}}\right| \right)} - 1.66 \log{\left(\left|{x_{6} - 2.852}\right| \right)} - 20.893$ \\ 
  & 7 & 0.687 & 0.69 & $0.001 x_{0}^{2} - 14.895 \log{\left(\left|{x_{0}}\right| \right)} - 0.044 \log{\left(\left|{x_{46} x_{48}}\right| \right)} + 10.243 \log{\left(\left|{x_{0} - 1.452 x_{59}}\right| \right)} + 2.696 \log{\left(\left|{x_{3} + x_{6}}\right| \right)} - 0.151 \log{\left(\left|{2.953 x_{36} + 0.025}\right| \right)} + 2.002 \log{\left(\frac{2.785}{\left|{\sqrt{x_{6}^{2} + 1.0}}\right|} \right)} - 10.89 - \frac{4.576}{x_{6}^{2} + 1.0}$ \\ 
\midrule 
\multirow{3}{*}{\BCR} & 3 & 0.62 & 0.624 & $- 0.001 x_{1}^{4} + 0.042 x_{3} + 0.051 \log{\left(\left|{x_{6}}\right| \right)}^{2} - 0.051 \log{\left(\left|{x_{6}}\right| \right)} - 0.138 \log{\left(\left|{\log{\left(\left|{x_{1}}\right| \right)}}\right| \right)} - 0.201 - \frac{5.841}{x_{1}^{2} + 1.0}$ \\ 
  & 5 & 0.629 & 0.629 & $- 0.179 x_{27} + 0.041 x_{3} - 1.629 \left(0.328 x_{1} - 1\right)^{2} - 0.015 \log{\left(\left|{x_{36}}\right| \right)} + 0.3 \log{\left(\left|{x_{6} - 3.77}\right| \right)} - 0.065 \log{\left(\left|{\log{\left(\left|{x_{1}}\right| \right)}}\right| \right)} - 0.026 \log{\left(\left|{\log{\left(\left|{x_{6}}\right| \right)}}\right| \right)} + 1.039$ \\ 
  & 7 & 0.641 & 0.616 & $- 0.221 x_{27} + 0.05 x_{3} - 0.43 x_{36}^{2} - 0.026 \log{\left(\left|{x_{20} x_{46}}\right| \right)} - 0.251 \log{\left(\left|{x_{1} - 3.205}\right| \right)} + 0.335 \log{\left(\left|{x_{6} - 4.07}\right| \right)} - 0.063 \log{\left(\left|{\log{\left(\left|{x_{1}}\right| \right)}}\right| \right)} - 0.024 \log{\left(\left|{\log{\left(\left|{x_{6}}\right| \right)}}\right| \right)} + 1.574$ \\

         \bottomrule
    \end{tabular}
\end{table*}

\subsection{Performance}
\label{sec:predictiveperformance}

\subsubsection{SR outperforms other glass-box methods}
\Cref{tab:hvtable} shows the test \HV of the Pareto front for the glass-box methods, at varying cutoff points in the front, i.e, taking the front filtered to contain only models with up to $k$ dimensions (``$\text{max}$'' indicates the whole front is taken).
CX consistently outperforms ST and, importantly, SR consistently outperforms both CX and ST across datasets and cutoff points $k$.
The only case in which CX beats SR is on normalized \SPP for Pareto fronts including models with more than $5$ dimensions.

We also find that no statistical significant differences are present when comparing SR with and without normalization (not shown).
ST, being tree-based, is normalization-agnostic but performs poorly in both with and without normalization.
Conversely, CX needs normalization to work well on some datasets, as can be seen by looking at its \HV scores on \PBC and \SPP in~\Cref{tab:hvtable}.
Similar findings are presented in~\Cref{tab:citable}, where we focus on the test \CI of models with exactly $k$ dimensions. 
SR delivers the most accurate models in the majority of cases, except for \SPP when we have $k \geq 5$ and for \SPP and \BCM when the maximally-dimensional models from the front are considered, in which case CX performs best.

\subsubsection{SR is competitive with black-box methods}
In~\Cref{tab:citableblackbox} we focus purely on predictive performance, and report comparisons between the highest-dimensional SR models and the black-box GB and RF models.
Since the black-box methods are normally run on normalized data, the table report results on the normalized datasets. We confirm that nearly-identical results are obtained without normalization (as mentioned in the previous sub-section regarding SR, while tree-based algorithms are invariant to numerical scale).
Here, SR is statistically significantly outperformed by GB and RF on \PBC and on \BCR, while it performs on par with GB and RF on the other datasets, and even significantly outperforms RF on \FRM.

\subsubsection{Qualitative visualizations}
In~\Cref{fig:paretoallmethodsplot}, we show the median Pareto front (in terms of test \HV) for each glass-box method, alongside with the median \CI for the black-box methods, for the normalized datasets.
The Pareto fronts of SR dominate those of CX and ST on \FRM, \BCM, and \BCR; interestingly, often with fewer and smaller-dimensional models. 
Moreover, in line with the findings of \Cref{tab:citableblackbox}, SR delivers models that compete with the black-box ones in terms of \CI (except on \PBC).

We manually inspect the survival functions (probability of survival over time) predicted by the methods, and show an example in~\Cref{fig:survfuncplot} for normalized \BCM.
Overall, we find that models behave similarly between SR and CX, and in turn these are not too dissimilar from GB and RF.
Conversely, ST stands out by predicting rather discretized survival functions, which are fairly different from those of the other methods.



\subsection{Readability and Interpretability}
\label{sec:readability}

\Cref{fig:multitreelengthlineplot} shows the relationship between the median number of expressions with respect to the dimensionality of the discovered models $k$.
The plots shows a correlation between number of expressions and $k$ (median Pearson: 0.81), while the number of expressions remains relatively contained (less than $10$).
We recall that each expression is limited in size due to imposed constraints (\Cref{sec:setting}).

\Cref{tab:gpformulae} shows examples of obtained expressions, i.e., $\bm \theta^\intercal \bm f$ in \Cref{eq:our-cox}, alongside their train and test \CI, from random repetitions (with no dataset normalization). 
Notably, evolution discovers expressions containing both linear and non-linear terms.
Arguably, the expressions are reasonably contained in size and dimensions, and stand a chance of being interpretable.
As this paper focuses on methodology, we do not assign meaning to the features and attempt to interpret the expressions here.
We note that our use of protected operations (AQ and ProtectedLog), which is intended to prevent numerical errors, likely complicates interpretation.
We also note cases of overfitting, where simpler expressions obtain higher \CI than higher dimensional ones.
For example on \BCM, the $3$-dimensional model generalizes better than the higher-dimensional ones.
This means that incorporating overfitting detection could lead to better generalizing and more interpretable models.

\section{Conclusion}
\label{sec:conclusion}

We propose an evolutionary Symbolic Regression (SR) algorithm adapted to survival regression to obtain accurate and interpretable models via multi-objective and multi-expression evolution.
Our experiments on five real-world datasets show that SR leads to more accurate survival than traditional regularized Cox proportional hazard models and survival trees.
Moreover, SR models also fare competitively with black-box gradient boosting survival and random survival forest methods. 
With qualitative results, we show that SR models stand a chance of being interpretable, even though the use of protected operations, aimed at preventing numerical errors in this paper, can be detrimental.
Overall, our work shows that SR can be a promising direction for survival regression. 
Future work should consider SR approaches that can overcome the proportional hazard assumption, the use of regularization and cross-validation techniques to prevent overfitting, and methods to deal with numerical errors without resorting to unprotected operations to improve interpretability.

\begin{acks}
We thank Dr.~Giuseppe Pasculli from InSilicoTrials Technologies for insightful early discussions.
\end{acks}

\bibliographystyle{ACM-Reference-Format}
\bibliography{refs}





\end{document}